\documentclass[11pt,letterpaper]{article}
\usepackage{emnlp2016}
\usepackage{times}
\usepackage{latexsym}

\emnlpfinalcopy

\def\emnlppaperid{***}

\usepackage[utf8]{inputenc} 
\usepackage[T1]{fontenc}    
\usepackage{hyperref}       
\usepackage{url}            
\usepackage{booktabs}       
\usepackage{amsfonts}       
\usepackage{nicefrac}       
\usepackage{microtype}      
\usepackage[centertags]{amsmath}
\usepackage{amssymb}
\usepackage{graphicx}

\def\x{\times}
\def\S{\mathbf{S}}
\def\H{\mathbf{H}}

\newcommand{\LL}{\mathcal{L}}

\newcommand{\QQ}{\mathcal{Q}}

\newcommand{\TT}{\mathcal{T}}

\DeclareMathOperator*{\expect}{\mathbb{E}}

\title{Natural Language Comprehension with the EpiReader}

\author{Adam Trischler \\ {\tt adam.trischler} \And Zheng Ye \\ {\tt jeff.ye} \And Xingdi Yuan \\ {\tt eric.yuan} \AND
        Kaheer Suleman \\ {\tt k.suleman@maluuba.com} \\ Maluuba Research \\ Montreal, Qu\'{e}bec, Canada }

\begin{document}

\maketitle

\begin{abstract}
  We present the EpiReader, a novel model for machine comprehension of text. Machine comprehension of unstructured, real-world text is a major research goal for natural language processing. Current tests of machine comprehension pose questions whose answers can be inferred from some supporting text, and evaluate a model's response to the questions. The EpiReader is an end-to-end neural model comprising two components: the first component proposes a small set of candidate answers after comparing a question to its supporting text, and the second component formulates hypotheses using the proposed candidates and the question, then reranks the hypotheses based on their estimated concordance with the supporting text. We present experiments demonstrating that the EpiReader sets a new state-of-the-art on the CNN and Children's Book Test machine comprehension benchmarks, outperforming previous neural models by a significant margin.
\end{abstract}

\section{Introduction}
When humans reason about the world, we tend to formulate a variety of hypotheses and counterfactuals, then test them in turn by physical or thought experiments. The philosopher Epicurus first formalized this idea in his Principle of Multiple Explanations: if several theories are consistent with the observed data, retain them all until more data is observed. In this paper, we argue that the same principle can be applied to machine comprehension of natural language. We propose a deep, end-to-end, neural comprehension model that we call the EpiReader.

Comprehension of natural language by machines, at a near-human level, is a prerequisite for an extremely broad class of useful applications of artificial intelligence. Indeed, most human knowledge is collected in the natural language of text. Machine comprehension (MC) has therefore garnered significant attention from the machine learning research community. Machine comprehension is typically evaluated by posing a set of questions based on a supporting text passage, then scoring a system's answers to those questions. We all took similar tests in school. Such tests are objectively gradable and may assess a range of abilities, from basic understanding to causal reasoning to inference~\cite{richardson2013}.

In the past year, two large-scale MC datasets have been released: the CNN/Daily Mail corpus, consisting of news articles from those outlets~\cite{hermann2015}, and the Children's Book Test (CBT), consisting of short excerpts from books available through Project Gutenberg~\cite{hill2015}. The size of these datasets (on the order of $10^5$ distinct questions) makes them amenable to data-intensive deep learning techniques. Both corpora use Cloze-style questions~\cite{taylor1953}, which are formulated by replacing a word or phrase in a given sentence with a placeholder token. The task is then to find the answer that ``fills in the blank''.

In tandem with these corpora, a host of neural machine comprehension models has been developed~\cite{weston2014,hermann2015,hill2015,kadlec2016,chen2016}. We compare the EpiReader to these earlier models through training and evaluation on the CNN and CBT datasets.\footnote{The CNN and Daily Mail datasets were released together and have the same form. The Daily Mail dataset is significantly larger, and our tests with this data are still in progress.}

The EpiReader factors into two components. The first component extracts a small set of potential answers based on a shallow comparison of the question with its supporting text; we call this the \emph{Extractor}. The second component reranks the proposed answers based on deeper semantic comparisons with the text; we call this the \emph{Reasoner}.
We can summarize this process as \textsl{Extract} $\rightarrow$ \textsl{Hypothesize} $\rightarrow$ \textsl{Test}\footnote{The Extractor performs extraction, while the Reasoner both hypothesizes and tests.}.
The semantic comparisons implemented by the Reasoner are based on the concept of {\it recognizing textual entailment} (RTE)~\cite{dagan2006}, also known as natural language inference. This process is computationally demanding. Thus, the Extractor serves the important function of filtering a large set of potential answers down to a small, tractable set of likely candidates for more thorough testing.

The Extractor follows the form of a pointer network~\cite{vinyals2015}, and uses a differentiable attention mechanism to indicate words in the text that potentially answer the question. This approach was used (on its own) for question answering with the Attention Sum Reader~\cite{kadlec2016}. The Extractor outputs a small set of answer candidates along with their estimated probabilities of correctness. The Reasoner forms hypotheses by inserting the candidate answers into the question, then estimates the concordance of each hypothesis with each sentence in the supporting text. We use these estimates as a measure of the evidence for a hypothesis, and aggregate evidence over all sentences. In the end, we combine the Reasoner's evidence with the Extractor's probability estimates to produce a final ranking of the answer candidates.

This paper is organized as follows. In Section~\ref{sec:probnote} we formally define the problem to be solved and give some background on the datasets used in our tests. In Section~\ref{sec:epireader} we describe the EpiReader, focusing on its two components and how they combine. Section~\ref{sec:related} discusses related work, and Section~\ref{sec:res} details our experimental results and analysis. We conclude in Section~\ref{sec:conc}.

\section{Problem definition, notation, datasets}
\label{sec:probnote}
The task of the EpiReader is to answer a Cloze-style question by reading and comprehending a supporting passage of text. The training and evaluation data consist of tuples $(\mathcal{Q}, \mathcal{T}, a^{\ast}, A)$, where $\mathcal{Q}$ is the question (a sequence of words $\{q_1,...q_{|\mathcal{Q}|}\}$), $\mathcal{T}$ is the text (a sequence of words $\{t_1,...,t_{|\mathcal{T}|}\}$), $A$ is a set of possible answers $\{a_1,...,a_{|A|}\}$, and $a^{\ast} \in A$ is the correct answer. All words come from a vocabulary $V$, and $A \subset \mathcal{T}$. In each question, there is a placeholder token indicating the missing word to be filled in.


\subsection{Datasets}
\paragraph{CNN} This corpus is built using articles scraped from the CNN website. The articles themselves form the text passages, and questions are generated synthetically from short summary statements that accompany each article. These summary points are (presumably) written by human authors. Each question is created by replacing a named entity in a summary point with a placeholder token. All named entities in the articles and questions are replaced with anonymized tokens that are shuffled for each $(\mathcal{Q}, \mathcal{T})$ pair. This forces the model to rely only on the text, rather than learning world knowledge about the entities during training. The CNN corpus (henceforth CNN) was presented by~\newcite{hermann2015}.

\paragraph{Children's Book Test} This corpus is constructed similarly to CNN, but from children's books available through Project Gutenberg. Rather than articles, the text passages come from book excerpts of 20 sentences. Since no summaries are provided, a question is generated by replacing a single word in the next (i.e.~21st) sentence. The corpus distinguishes questions based on the type of word that is replaced: named entity, common noun, verb, or preposition. Like~\newcite{kadlec2016}, we focus only on the first two classes since~\newcite{hill2015} showed that standard LSTM language models already achieve human-level performance on the latter two. Unlike in the CNN corpora, named entities are not anonymized and shuffled in the Children's Book Test (CBT). The CBT was presented by~\newcite{hill2015}.

\paragraph{} Due to the construction of questions in each corpus, CNN and CBT assess different aspects of machine comprehension. The summary points of CNN are condensed paraphrasings of information from the text, so determining the correct answer relies more on recognizing textual entailment. On the other hand, a CBT question, generated from a sentence which continues the text passage (rather than summarizes it), is more about making a prediction based on context.

\section{The EpiReader}
\label{sec:epireader}
\subsection{Overview and intuition}
The EpiReader explicitly leverages the observation that the answer to a question is often a word or phrase from the related text passage. This condition holds for the CNN and CBT datasets. EpiReader's first module, the Extractor, can thus select a small set of candidate answers by pointing to their locations in the supporting passage. This mechanism is detailed in Section~\ref{sec:extractor}, and was used previously by the Attention Sum Reader~\cite{kadlec2016}. Pointing to candidate answers removes the need to apply a softmax over the entire vocabulary as in~\newcite{weston2014}, which is computationally more costly and uses less-direct information about the context of a predicted answer in the supporting text.

EpiReader's second module, the Reasoner, begins by formulating hypotheses using the extracted answer candidates. It generates each hypothesis by replacing the placeholder token in the question with an answer candidate. Cloze-style questions are ideally-suited to this process, because inserting the correct answer at the placeholder location produces a well-formed, grammatical statement. Thus, the correct hypothesis will ``make sense'' to a language model.

\begin{figure*}[t]
  \centering
  \includegraphics[width=5in]{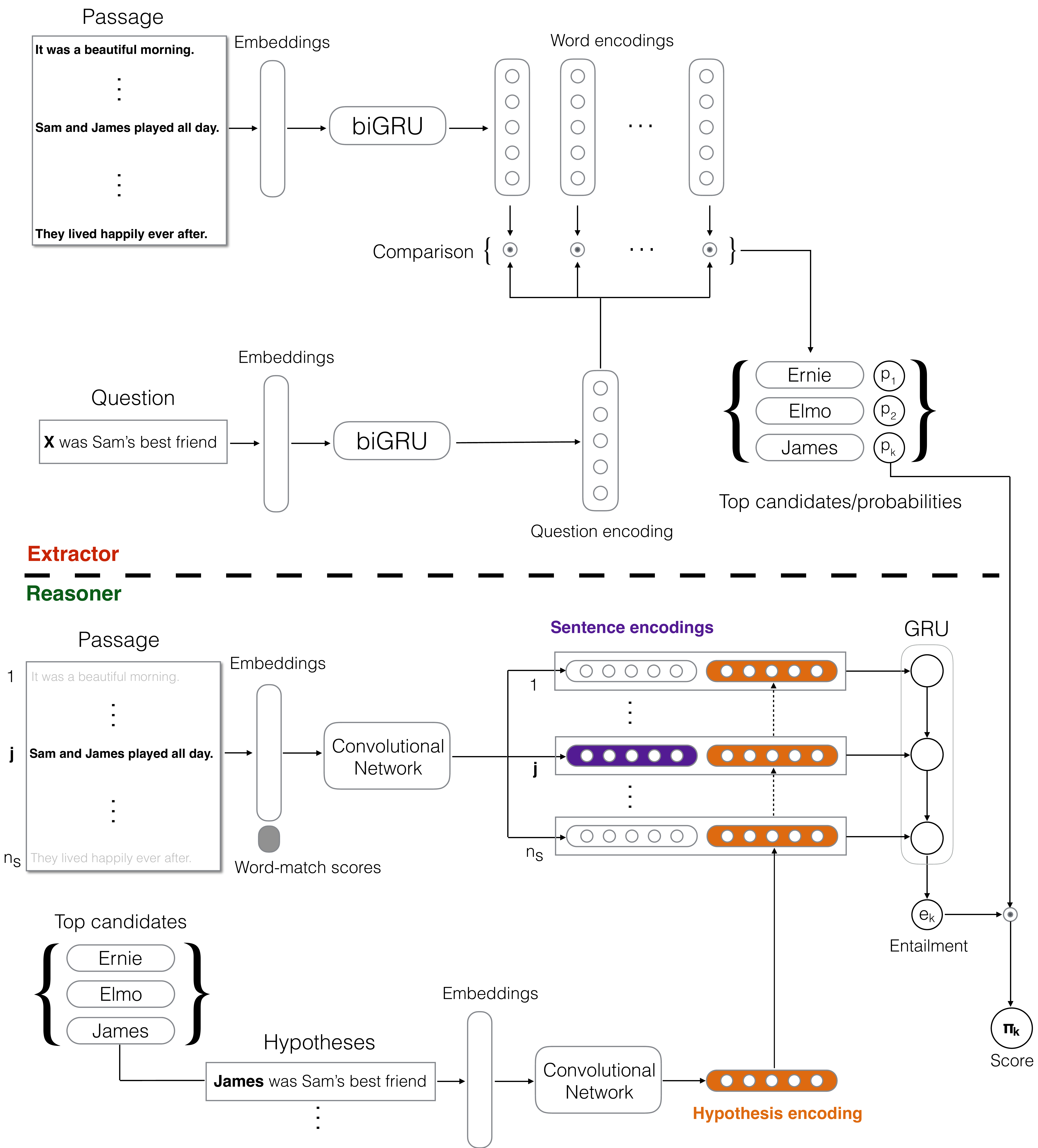}
  \caption{The complete EpiReader framework. The Extractor is above, the Reasoner below. Propagating the Extractor's probability estimates forward and combining them with the Reasoner's entailment estimates renders the model end-to-end differentiable.}
  \label{fig:framework}
\end{figure*}

The Reasoner then tests each hypothesis individually. It compares a hypothesis to the text, split into sentences, to measure textual entailment, and then aggregates entailment over all sentences. This computation uses a pair of convolutional encoder networks followed by a recurrent neural network. The convolutional encoders generate abstract representations of the hypothesis and each text sentence; the recurrent network estimates and aggregates entailment. This is described formally in Section~\ref{sec:reasoner}. The end-to-end EpiReader model, combining the Extractor and Reasoner modules, is depicted in Figure~\ref{fig:framework}.

Throughout our model, words will be represented with trainable embeddings~\cite{bengio2000}. We represent these embeddings using a matrix $\mathbf{W} \in \mathbb{R}^{D \x |V|}$, where $D$ is the embedding dimension and $|V|$ is the vocabulary size.

\subsection{The Extractor}
\label{sec:extractor}
The Extractor is a Pointer Network \cite{vinyals2015}. It uses a pair of bidirectional recurrent neural networks, $f(\theta_T, \mathbf{T})$ and $g(\theta_Q, \mathbf{Q})$, to encode the text passage and the question. $\theta_T$ represents the parameters of the text encoder, and $\mathbf{T} \in \mathbb{R}^{D \x N}$ is a matrix representation of the text (comprising $N$ words), whose columns are individual word embeddings $\mathbf{t}_i$. Likewise, $\theta_Q$ represents the parameters of the question encoder, and $\mathbf{Q} \in \mathbb{R}^{D \x N_Q}$ is a matrix representation of the question (comprising $N_Q$ words), whose columns are individual word embeddings $\mathbf{q}_j$.

We use a recurrent neural network with gated recurrent units (GRU)~\cite{bahdanau2014} to scan over the columns (i.e.~word embeddings) of the input matrix.
We selected the GRU because it is computationally simpler than Long Short-Term Memory~\cite{hochreiter1997}, while still avoiding the problem of vanishing/exploding gradients often encountered when training recurrent networks.

The GRU's hidden state gives a representation of the $i$th word conditioned on preceding words. To include context from proceeding words, we run a second GRU over $\mathbf{T}$ in the reverse direction. We refer to the combination as a biGRU. At each step the biGRU outputs two $d$-dimensional encoding vectors, one for the forward direction and one for the backward direction. We concatenate these to yield a vector $f(\mathbf{t}_i) \in \mathbb{R}^{2d}$. The question biGRU is similar, but we get a single-vector representation of the question by concatenating the final forward state with the initial backward state, which we denote $g(\mathbf{Q}) \in \mathbb{R}^{2d}$.

As in~\newcite{kadlec2016}, we model the probability that the $i$th word in text $\mathcal{T}$ answers question $\mathcal{Q}$ using
\begin{equation}
	s_i \propto \exp(f(\mathbf{t}_i) \cdot g(\mathbf{Q})),
	\label{eq:k-prob}
\end{equation}
which takes the inner product of the text and question representations followed by a softmax. In many cases unique words repeat in a text. Therefore, we compute the total probability that word $w$ is the correct answer using a sum:
\begin{equation}
	P(w \, | \, \mathcal{T}, \mathcal{Q}) = \sum_{i:~t_i=w} s_i.
	\label{eq:prob-sum}
\end{equation}
This probability is evaluated for each unique word in $\mathcal{T}$. Finally, the Extractor outputs the set $\{p_1, ..., p_K\}$ of the $K$ highest word probabilities from~\ref{eq:prob-sum}, along with the corresponding set of $K$ most probable answer words $\{\hat{a}_1, ..., \hat{a}_K \}$.

\subsection{The Reasoner}
\label{sec:reasoner}
The indicial selection involved in gathering $\{\hat{a}_1, ..., \hat{a}_K \}$ is not a smooth operation. To construct an end-to-end differentiable model, we bypass this by propagating the probability estimates of the Extractor directly through the Reasoner.

The Reasoner begins by inserting the answer candidates, which are single words or phrases, into the question sequence $\mathcal{Q}$ at the placeholder location. This forms $K$ hypotheses $\{\mathcal{H}_1, ..., \mathcal{H}_K\}$. At this point, we consider each hypothesis to have probability $p(\mathcal{H}_k) \approx p_k$, as estimated by the Extractor. The Reasoner updates and refines this estimate.

The hypotheses represent new information in some sense---they are statements we have constructed, albeit from words already present in the question and text passage. The Reasoner estimates entailment between the statements $\mathcal{H}_k$ and the passage $\mathcal{T}$. We denote these estimates using $e_k = F(\mathcal{H}_k, \mathcal{T})$, with $F$ to be defined. We start by reorganizing $\mathcal{T}$ into a sequence of $N_s$ sentences: $\mathcal{T} = \{t_1, \ldots, t_N\} \rightarrow \{\mathcal{S}_1, \ldots, \mathcal{S}_{N_s}\}$, where $\mathcal{S}_i$ is a sequence of words.

For each hypothesis and each sentence of the text, Reasoner input consists of two matrices: $\S_i \in \mathbb{R}^{D \x |\mathcal{S}_i|}$, whose columns are the embedding vectors for each word of sentence $\mathcal{S}_i$, and $\H_k \in \mathbb{R}^{D \x |\mathcal{H}_k|}$, whose columns are the embedding vectors for each word in the hypothesis $\mathcal{H}_k$. The embedding vectors themselves come from matrix $\mathbf{W}$, as before.

These matrices feed into a convolutional architecture based on that of~\newcite{severyn2016}. The architecture first augments $\S_i$ with matrix $\mathbf{M} \in \mathbb{R}^{2 \x |\mathcal{S}_i|}$. The first row of $\mathbf{M}$ contains the inner product of each word embedding in the sentence with the candidate answer embedding, and the second row contains the maximum inner product of each sentence word embedding with any word embedding in the question. These word-matching features were inspired by similar approaches in~\newcite{wang2015} and~\newcite{trischler2016}, where they were shown to improve entailment estimates.

The augmented $\S_i$ is then convolved with a bank of filters $\mathbf{F}^S \in \mathbb{R}^{(D+2) \x m}$, while $\H_k$ is convolved with filters $\mathbf{F}^H \in \mathbb{R}^{D \x m}$, where $m$ is the convolutional filter width. We add a bias term and apply a nonlinearity (we use a ReLU) following the convolution. Maxpooling over the sequences then yields two vectors: the representation of the text sentence, $\mathbf{r}_{\mathcal{S}_i} \in \mathbb{R}^{N_F}$, and the representation of the hypothesis, $\mathbf{r}_{\mathcal{H}_k} \in \mathbb{R}^{N_F}$, where $N_F$ is the number of filters.

We then compute a scalar similarity score between these vector representations using the bilinear form
\begin{equation}
	\varsigma = \mathbf{r}^T_{\mathcal{S}_i} \mathbf{R}  \mathbf{r}_{\mathcal{H}_k},
	\label{eq:simscore}
\end{equation}
where $\mathbf{R} \in \mathbb{R}^{N_F \x N_F}$ is a matrix of trainable parameters. We then concatenate the similarity score with the sentence and hypothesis representations to get a vector, $\mathbf{x}_{ik} = [\varsigma; \mathbf{r}_{\mathcal{S}_i}; \mathbf{r}_{\mathcal{H}_k}]^T$. There are more powerful models of textual entailment that could have been used in place of this convolutional architecture. We adopted the approach of \newcite{severyn2016} for computational efficiency.

The resulting sequence of $N_s$ vectors feeds into yet another GRU for synthesis, of hidden dimension $d_S$. Intuitively, it is often the case that evidence for a particular hypothesis is distributed over several sentences. For instance, if we hypothesize that \textsl{the football is in the park}, perhaps it is because one sentence tells us that \textsl{Sam picked up the football} and a later one tells us that \textsl{Sam ran to the park}.\footnote{This example is characteristic of the {\it bAbI} dataset~\cite{weston2015}.} The Reasoner synthesizes distributed information by running a GRU network over $\mathbf{x}_{ik}$, where $i$ indexes sentences and represents the step dimension.\footnote{Note a benefit of forming the hypothesis: it renders bidirectional aggregation unnecessary, since knowing both the question and the putative answer "closes the loop" the same way that a bidirectional encoding would.} The final hidden state of the GRU is fed through a fully-connected layer, yielding a single scalar $y_k$. This value represents the collected evidence for $\mathcal{H}_k$ based on the text. In practice, the Reasoner processes all $K$ hypotheses in parallel and the estimated entailment of each is normalized by a softmax, $e_k \propto \exp(y_k)$.

The reranking step performed by the Reasoner helps mitigate a significant weakness of most existing attention mechanisms. Specifically, these mechanisms blend representations of all possible outcomes together using ``soft'' attention, rather than considering them discretely using ``hard'' attention. This is like exploring a maze by generating an average path out of the several before you, and then attempting to follow it by walking through a wall. Examining possibilities individually, as in the Reasoner module, is more natural.

\subsection{Combining components}
Finally, we combine the evidence from the Reasoner with the probability from the Extractor.
We compute the output probability of each hypothesis, $\pi_k$, according to the product
\begin{equation}
	\pi_k \propto e_k p_k,
	\label{eq:bayes}
\end{equation}
whereby the evidence of the Reasoner can be interpreted as a correction to the Extractor probabilities, applied as an additive shift in $\log$-space.
We experimented with other combinations of the Extractor and Reasoner, but we found the multiplicative approach to yield the best performance. 

After combining results from the Extractor and Reasoner to get the probabilities $\pi_k$ described in Eq.~\ref{eq:bayes}, we optimize the parameters of the full EpiReader to minimize a cost comprising two terms, $\LL_{E}$ and $\LL_{R}$. The first term is a standard negative log-likelihood objective, which encourages the Extractor to rate the correct answer above other answers. This is the same loss term used in \newcite{kadlec2016}. This loss is given by:
\begin{equation}
\mathcal{L}_{E} = \expect_{(\QQ, \TT, a^{\ast}, A)}\left[ - \log P(a^{\ast} \, | \, \TT, \QQ) \right],
\label{eq:extractor_loss}
\end{equation}
where $P(a^{\ast} \, | \, \TT, \QQ)$ is as defined in Eq.~\ref{eq:prob-sum}, and $a^{\ast}$ denotes the true answer. The second term is a margin-based loss on the end-to-end probabilities $\pi_k$. We define $\pi^{\ast}$ as the probability $\pi_k$ corresponding to the true answer word $a^{\ast}$. This term is given by:
\begin{equation}
\mathcal{L}_{R} = \expect_{(\QQ, \TT, a^{\ast}, A)} \left[ \sum_{\hat{a}_i \in \{\hat{a}_1, ..., \hat{a}_K\} \setminus a^{\ast}} [\gamma - \pi^{\ast} + \pi_{\hat{a}_i}]_{+}  \right],
\label{eq:reasoner_loss}
\end{equation}
where $\gamma$ is a margin hyperparameter, $\{\hat{a}_1, ..., \hat{a}_K\}$ is the set of $K$ answers proposed by the Extractor, and $[ x ]_{+}$ indicates truncating $x$ to be non-negative. Intuitively, this loss says that we want the end-to-end probability $\pi^{\ast}$ for the correct answer to be at least $\gamma$ larger than the probability $\pi_{\hat{a}_i}$ for any other answer proposed by the Extractor. During training, the correct answer is occasionally missed by the Extractor, especially in early epochs. We counter this issue by forcing the correct answer into the top $K$ set while training. When evaluating the model on validation and test examples we rely fully on the top $K$ answers proposed by the Extractor.

To get the final loss term $\LL_{ER}$, minus $\ell_2$ regularization terms on the model parameters, we take a weighted combination of $\LL_{E}$ and $\LL_{R}$:
\begin{equation}
\LL_{ER} = \LL_{E} + \lambda \LL_{R},
\label{eq:joint_loss}
\end{equation}
where $\lambda$ is a hyperparameter for weighting the relative contribution of the Extractor and Reasoner losses. In practice, we found that $\lambda$ should be fairly large (e.g.~$10 < \lambda < 100$).
Empirically, we observed that the output probabilities from the Extractor often peak and saturate the first softmax; hence, the Extractor term can come to dominate the Reasoner term without the weight $\lambda$ (we discuss the Extractor's propensity to overfit in Section~\ref{sec:res}).

\section{Related Work}
\label{sec:related}
The Impatient and Attentive Reader models were proposed by~\newcite{hermann2015}. The Attentive Reader applies bidirectional recurrent encoders to the question and supporting text. It then uses the attention mechanism described in~\newcite{bahdanau2014} to compute a fixed-length representation of the text based on a weighted sum of the text encoder's output, guided by comparing the question representation to each location in the text. Finally, a joint representation of the question and supporting text is formed by passing their separate representations through a feedforward MLP and an answer is selected by comparing the MLP output to a representation of each possible answer. The Impatient Reader operates similarly, but computes attention over the text after processing each consecutive word of the question. The two models achieved similar performance on the CNN and Daily Mail datasets.

Memory Networks were first proposed by~\newcite{weston2014} and later applied to machine comprehension by~\newcite{hill2015}. This model builds fixed-length representations of the question and of windows of text surrounding each candidate answer, then uses a weighted-sum attention mechanism to combine the window representations. As in the previous Readers, the combined window representation is then compared with each possible answer to form a prediction about the best answer. What distinguishes Memory Networks is how they construct the question and text window representations. Rather than a recurrent network, they use a specially-designed, trainable transformation of the word embeddings.

Most of the details for the very recent AS Reader are provided in the description of our Extractor module in Section~\ref{sec:extractor}, so we do not summarize it further here. This model~\cite{kadlec2016} set the previous state-of-the-art on the CBT dataset.

During the write-up of this paper, another very recent model came to our attention. \newcite{chen2016} propose using a bilinear term instead of a $\tanh$ layer to compute the attention between question and passage words, and also uses the attended word encodings for direct, pointer-style prediction as in~\newcite{kadlec2016}. This model set the previous state-of-the-art on the CNN dataset. However, this model used embedding vectors pretrained on a large external corpus~\cite{pennington2014}.

The EpiReader borrows ideas from other models as well. The Reasoner's convolutional architecture is based on~\newcite{severyn2016} and~\newcite{kalchbrenner2014}. Our use of word-level matching was inspired by the Parallel-Hierarchical model of~\newcite{trischler2016} and the natural language inference model of~\newcite{wang2015}. Finally, the idea of formulating and testing hypotheses for question-answering was used to great effect in IBM's DeepQA system for {\it Jeopardy!}~\cite{ferrucci2010}, although that was a more traditional information retrieval pipeline rather than an end-to-end neural model.

\section{Evaluation}
\label{sec:res}
\subsection{Implementation and training details}
To train our model we used stochastic gradient descent with the ADAM optimizer~\cite{kingma2014}, with an initial learning rate of 0.001. The word embeddings were initialized randomly, drawing from the uniform distribution over $[-0.05, 0.05)$. We used batches of 32 examples, and early stopping with a patience of 2 epochs. Our model was implement in Theano~\cite{theano10} using the Keras framework~\cite{keras}.

The results presented below for the EpiReader were obtained by searching over a small grid of hyperparameter settings. We selected the model that, on each dataset, maximized accuracy on the validation set, then evaluated it on the test set. We record the best settings for each dataset in Table~\ref{tab:hyper}. As has been done previously, we train separate models on CBT's named entity (CBT-NE) and common noun (CBT-CN) splits. All our models used $\ell_2$-regularization at $0.001$, $\lambda=50$, and $\gamma=0.04$. We did not use dropout but plan to investigate its effect in the future.
\begin{table}
  \caption{Hyperparameter settings for best EpiReaders. $D$ is the embedding dimension, $d$ is the hidden dimension in the Extractor GRUs, $K$ is the number of candidates to consider, $m$ is the filter width, $N_F$ is the number of filters, and $d_S$ is the hidden dimension in the Reasoner GRU.}
  \label{tab:hyper}
  \small
  \centering
  \begin{tabular}{lcccccc}
    \toprule
    {} & \multicolumn{6}{c}{Hyperparameters}          \\
 	\cmidrule{2-7}
    Dataset & $D$ & $d$ & $K$ & $m$ & $N_F$ & $d_S$            \\
    \midrule
    CBT-NE & 300 & 128 & 5 & 3 & 16 & 32 \\
    \midrule
    CBT-CN & 300 & 128 & 5 & 3 & 32 & 32 \\
    \midrule
    CNN & 384 & 256 & 10 & 3 & 32 & 32 \\ 
    \bottomrule
  \end{tabular}
\end{table}
\newcite{hill2015} and~\newcite{kadlec2016} also present results for ensembles of their models. Time did not permit us to generate an ensemble of EpiReaders on the CNN dataset so we omit those measures; however, EpiReader ensembles (of seven models) demonstrated improved performance on the CBT dataset.

\subsection{Results}
In Table~\ref{tab:cbt}, we compare the performance of the EpiReader against that of several baselines, on the validation and test sets of the CBT and CNN corpora. We measure EpiReader performance at the output of both the Extractor and the Reasoner.
\begin{table*}[t]
	\caption{Model comparison on the CBT and CNN datasets. Results marked with $^1$ are from Hill et al. (2016), those marked with $^2$ are from Kadlec et al. (2016), those marked with $^3$ are from Hermann et al. (2015), and those marked with $^4$ are from Chen et al. (2016).
	}
  \label{tab:cbt}
  \small
  \centering
  \begin{tabular}{lcccc}
    \toprule
    {} & \multicolumn{2}{c}{CBT-NE}   &    \multicolumn{2}{c}{CBT-CN}          \\
 	\cmidrule{2-3} \cmidrule{4-5}
    Model & valid & test & valid & test             \\
    \midrule
    Humans (context + query) $^1$ & - & 81.6 & - & 81.6 \\
    \midrule
    LSTMs (context + query) $^1$ & 51.2 & 41.8 & 62.6 & 56.0 \\
    \midrule
    MemNNs $^1$  & 70.4 & 66.6 & 64.2 & 63.0 \\
    \midrule
    AS Reader $^2$ & 73.8 & 68.6 & 68.8 & 63.4 \\
    \midrule
    EpiReader Extractor & 73.2 & 69.4 & 69.9 & 66.7 \\
    EpiReader & \textbf{75.3} & \textbf{69.7} & \textbf{71.5} & \textbf{67.4} \\
    \midrule \midrule
    AS Reader (ensemble) $^2$ & 74.5 & 70.6 & 71.1 & 68.9 \\
    EpiReader (ensemble) & \textbf{76.6} & \textbf{71.8} & \textbf{73.6} & \textbf{70.6} \\
    \bottomrule
  \end{tabular}
  \quad
  \begin{tabular}{lcc}
    \toprule
    {} & \multicolumn{2}{c}{CNN} \\
 	\cmidrule{2-3}
    Model & valid & test            \\
    \midrule
    Deep LSTM Reader $^3$ & 55.0 & 57.0 \\
    Attentive Reader $^3$ & 61.6 & 63.0 \\
    Impatient Reader $^3$ & 61.8 & 63.8 \\
    \midrule
    MemNNs $^1$  & 63.4 & 66.8 \\
    \midrule
    AS Reader $^2$ & 68.6 & 69.5 \\
    \midrule
    Stanford AR $^4$ & 72.4 & 72.4 \\
    \midrule
    EpiReader Extractor & 71.8 & 72.0 \\
    EpiReader & \textbf{73.4} & \textbf{74.0} \\
    \bottomrule
  \end{tabular}
\end{table*}
The EpiReader achieves state-of-the-art performance across the board for both datasets. On CNN, we score 2.2\% higher on test than the best previous model of~\newcite{chen2016}. Interestingly, an analysis of the CNN dataset by~\newcite{chen2016} suggests that approximately 25\% of the test examples contain coreference errors or questions which are ``ambiguous/hard'' even for a human analyst. If this estimate is accurate, then the EpiReader, achieving an absolute test accuracy of 74.0\%, is operating close to expected human performance. On the other hand, ambiguity is unlikely to be distributed evenly over entities, so a good model should be able to perform at better-than-chance levels even on questions where the correct answer is uncertain. If, on the 25\% of ``noisy'' questions, the model can shift its hit rate from, {\it e.g.}, 1/10 to 1/3, then there is still a fair amount of performance to gain.

On CBT-CN our single model scores 4.0\% higher than the previous best of the AS Reader. The improvement on CBT-NE is more modest at 1.1\%. Looking more closely at our CBT-NE results, we found that the validation and test accuracies had relatively high variance even in late epochs of training. We discovered that many of the validation and test questions were asked about the same named entity, which may explain this issue.

\subsection{Analysis}
Aside from achieving state-of-the-art results at its final output, the EpiReader framework gives a boost to its Extractor component through the joint training process. In Table~\ref{tab:cbt}, we provide accuracy scores evaluated at the output of the Extractor. These are all higher than the analogous scores reported for the AS Reader, which we verified ourselves to within negligible difference. Based on our own work with that model, we found it to overfit the training set rapidly and significantly, achieving training accuracy scores upwards of 98\% after only 2 epochs. We suspect that the Reasoner module had a regularizing effect on the Extractor, but leave the verification for future work. An analysis by~\newcite{kadlec2016} indicates that the trained AS Reader includes the correct answer among its five most probable candidates on approximately 95\% of test examples for both datasets. We verified that our Extractor achieved a similar rate, and of course this is vital for performance of the full system, since the Reasoner cannot recover when the correct answer is not among its inputs. 

\begin{figure}
  \centering
  \fbox{\includegraphics[width=\linewidth]{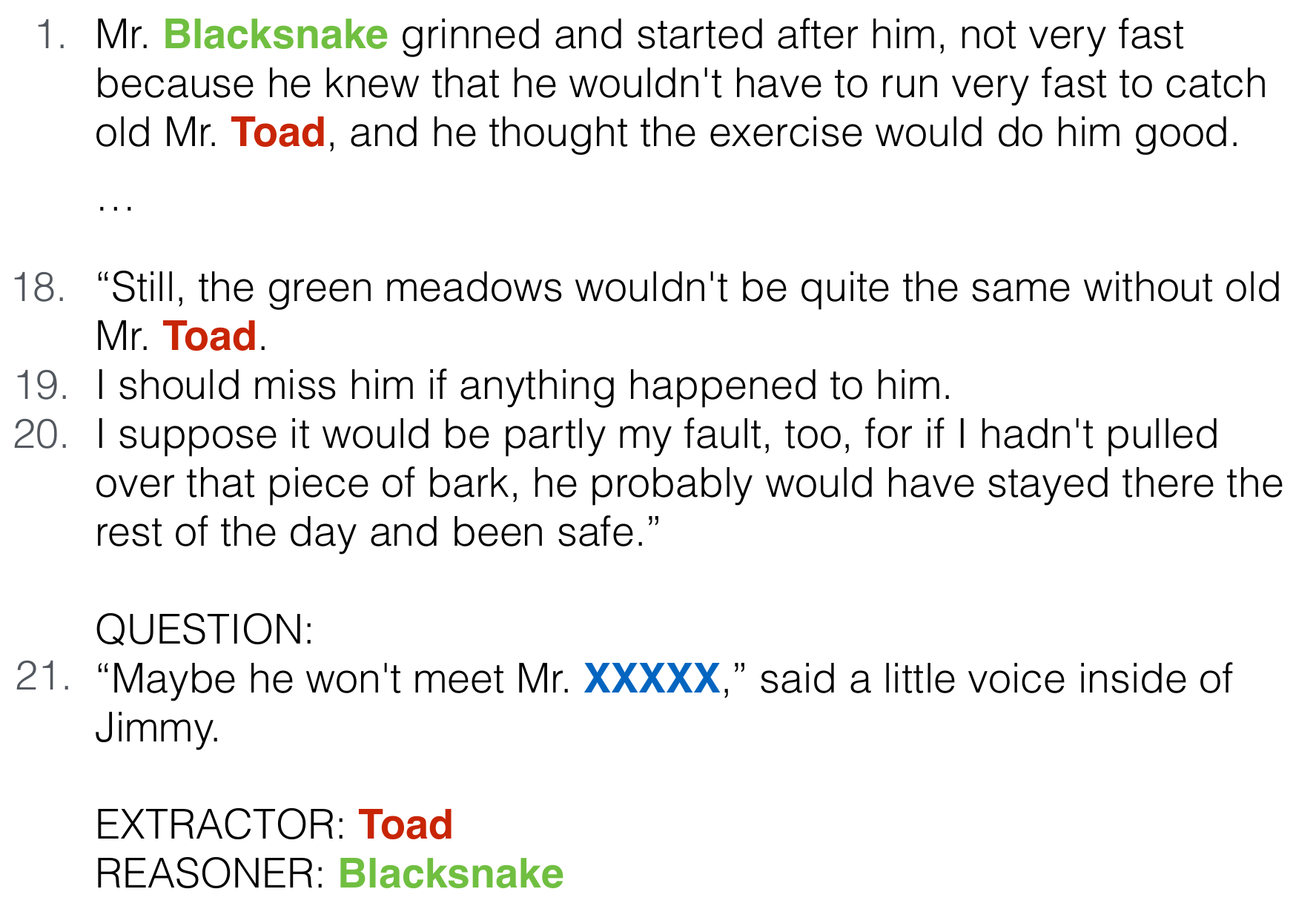}}
  \caption{An abridged example from CBT-NE demonstrating corrective reranking by the Reasoner.}
  \label{fig:snaketoad}
\end{figure}

Our results demonstrate that the Reasoner often corrects erroneous answers from the Extractor. Figure~\ref{fig:snaketoad} gives an example of this correction. In the text passage, from CBT-NE, Mr. Blacksnake is pursuing Mr. Toad, presumably to eat him. The dialogue in the question sentence refers to both: Mr. Toad is its subject, referred to by the pronoun ``he'', and Mr. Blacksnake is its object. In the preceding sentences, it is clear (to a human) that Jimmy is worried about Mr. Toad and his potential encounter with Mr. Blacksnake. The Extractor, however, points most strongly to ``Toad'', possibly because he has been referred to most recently. The Reasoner corrects this error and selects ``Blacksnake'' as the answer. This relies on a deeper understanding of the text. The named entity can, in this case, be inferred through an alternation of the entities most recently referred to. This kind alternation is typical of dialogues, when two actors interact in turns. The Reasoner can capture this behavior because it examines sentences in sequence.

\section{Conclusion}
\label{sec:conc}
In this article we presented the novel EpiReader framework for machine comprehension, and evaluated it on two large, complex datasets: CNN and CBT. Our model achieves state-of-the-art results on these corpora, outperforming all previous approaches. In future work, we plan to augment our framework with a more powerful model for natural language inference, and explore the effect of pretraining such a model specifically on an inference task. We also plan to try simplifying the model by reusing the Extractor's biGRU encodings in the Reasoner.

\bibliographystyle{emnlp2016}
\bibliography{nipsrefs}

\end{document}